\documentclass[english,letterpaper,10pt,conference]{ieeeconf}
\usepackage[T1]{fontenc}
\usepackage[latin9]{inputenc}
\usepackage{float}
\usepackage{amsmath}
\usepackage{graphicx}
\usepackage[numbers]{natbib}
\usepackage{tabularx}
\usepackage{multirow}
\usepackage{hyperref}

\makeatletter

\IEEEoverridecommandlockouts
\usepackage{url}
\usepackage{indentfirst}

\usepackage[footnotesize]{caption}

\usepackage{balance}
\usepackage{algorithm,algpseudocode}
\usepackage{color}

\@ifundefined{showcaptionsetup}{}{%
 \PassOptionsToPackage{caption=false}{subfig}}
\usepackage{subfig}
\makeatother

\usepackage{babel}
\begin{document}
\title{\textbf{SONIC: Sonar Image Correspondence using Pose Supervised Learning for Imaging Sonars}}
\author{Samiran Gode$^{*}$, Akshay Hinduja$^{*}$, and Michael Kaess\thanks{The authors are affiliated with the Robotics Institute, Carnegie Mellon University, Pittsburgh, PA 15213, USA. $^{*}$These authors contributed equally to this work. {\tt\small \{ahinduja, kaess\}@andrew.cmu.edu}, \tt\small{sgode}@alumni.cmu.edu}
\thanks{This work was partially supported by the Office of Naval Research award N00014-21-1-2482. The authors would also like to thank Eric Westman for providing code excerpts used in evaluation.}
}

\maketitle
\begin{abstract}
In this paper, we address the challenging problem of data association for underwater SLAM through a novel method for sonar image correspondence using learned features.  We introduce SONIC (SONar Image Correspondence), a pose-supervised network designed to yield robust feature correspondence capable of withstanding viewpoint variations. The inherent complexity of the underwater environment stems from the dynamic and frequently limited visibility conditions, restricting vision to a few meters of often featureless expanses. This makes camera-based systems suboptimal in most open water application scenarios.  Consequently, multibeam imaging sonars emerge as the preferred choice for perception sensors. However, they too are not without their limitations. While imaging sonars offer superior long-range visibility compared to cameras, their measurements can appear different from varying viewpoints. This inherent variability presents formidable challenges in data association, particularly for feature-based methods. Our method demonstrates significantly better performance in generating correspondences for sonar images which will pave the way for more accurate loop closure constraints and sonar-based place recognition. Code as well as simulated and real-world datasets are made public on \href{https://github.com/rpl-cmu/sonic}{https://github.com/rpl-cmu/sonic} to facilitate further development in the field. 
\end{abstract}

\begin{figure}[h!]\centering
\includegraphics[width=1\columnwidth]{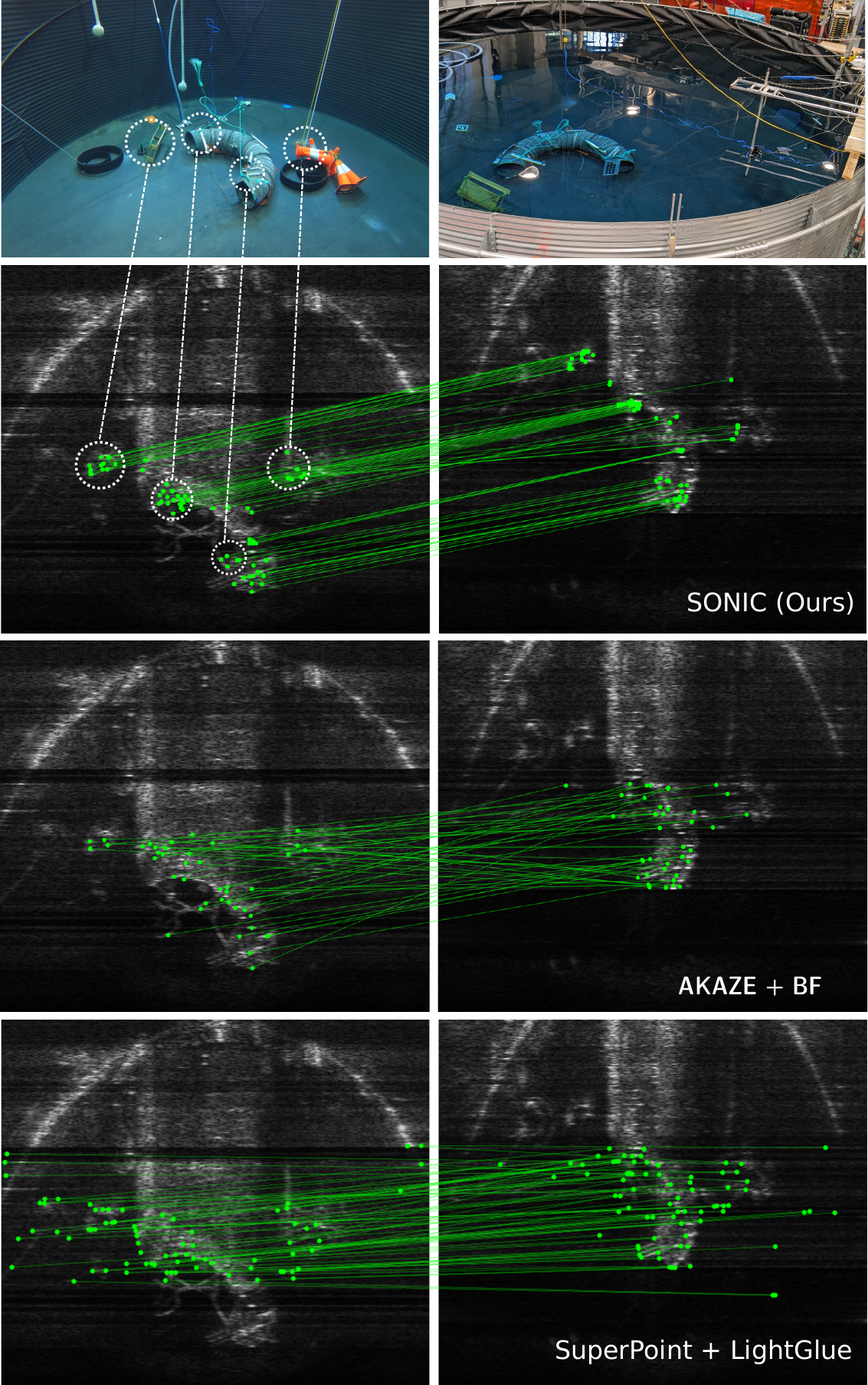}
\caption{Real-world matching performance: Sonar images taken from different planar positions in a test tank show our method providing significantly better matches than AKAZE with the brute force matcher, and the SuperPoint keypoints matched using LightGlue. Given the keypoints in the first frame, SONIC uses expectation matching to determine the correspondences and presents only those correspondences with high confidence. \vspace{-0.75cm}  \label{fig:match_performance}}
\end{figure} 
\section{Introduction \label{sec:Introduction-and-Related}}

Feature-based methods excel at localization and mapping by tracking distinct features over time. With camera frameworks, these methods utilize photometric consistency and invariance to common transformations such as scale and rotation, to produce precise feature correspondences~\cite{ORBSLAM2015}.

Camera-based localization and mapping face hurdles underwater due to reduced color depth and visibility, limiting usable data. In contrast, \textit{forward looking} or \textit{imaging} sonars excel underwater, offering long-range visibility impervious to water particulates. While sonars are optimal for such scenarios, they currently lack robust feature descriptors.

In the camera imaging domain, feature descriptors and detection are well-researched, with notable examples being  SIFT~\cite{zhangSIFT}, ORB~\cite{rublee2011orb} and AKAZE~\cite{alcantarilla2011fast}. 
The robustness of these feature descriptors stems mainly from two principles, photometric consistency and geometric invariance. Photometric consistency maintains pixel value stability against viewing angle variations and noise, while geometric resilience ensures feature recognition across varying orientations and scales.
On the other hand, imaging sonars exhibit variations in intensity returns and observable shapes when viewing the same object from different angles, influenced by the object's material and geometry.
These aforementioned descriptors struggle with the prevalent speckle noise and intensity variations characteristic of sonar images, especially in the polar space. An example of such a failure shown in Fig.~\ref{fig:match_performance}, in row 3 with AKAZE and the brute force (BF) matcher. The downstream effects of this failure can result in poor, or error-prone loop closure detection and state estimation. 
Our work addresses this issue by using pose supervision grounded in sonar geometry. This offers a feature correspondence method for sonar images, robust against photometric and geometric variations.
Recent research has improved upon the matching performance of camera images with deep neural network based approaches using ground truth correspondences~\cite{sarlin2020superglue,detone2018superpoint,sun2021loftr}. There has also been recent semi-supervised work techniques leveraging pose-supervised learning~\cite{wang2020learning}. These models, though powerful, are trained on camera data and suffer from the same problems explained above when used on sonar data. 

Motivated by~\cite{wang2020learning} and the availability of pose-supplemented sonar data using simulation \cite{potokar2022holoocean}, we propose a novel pose-supervised method to learn sonar image matching using a sonar epipolar-contour(see \ref{subsec:Sonar-Epipolar}) based loss function. In addition, we also enforce cyclic consistency as done in ~\cite{wang2020learning}. Our network learns directly from sonar images in the polar space and the loss is calculated in this space ensuring the model learns unique representations in sonar images. 
Specifically, our main contributions are:
\begin{itemize}
    \item  A novel semi-supervised method for sonar image correspondence, utilizing sonar-specific epipolar geometry negating the need for ground truth correspondences. This technique offers correspondences tailored for imaging sonar, producing features resilient to viewpoint changes.
    \item An extensive simulated dataset that comprises over 300K pairs of sonar images and their corresponding ground truth poses, collected across 10 distinct scenes. Each scene is made from randomized positioning of a specific set of objects, with repeatable sensor motion from different pose offsets. 
\end{itemize}


\section{Related Work} ~\label{sec:Related}

The descriptors mentioned in Section~\ref{sec:Introduction-and-Related} have been utilized for sonar images in the past for applications towards acoustic structure from motion (ASFM) and feature-based SLAM~\cite{Huang15iros,Westman18icra,Shin15oceans, Li18ral, Westman19joe}. These descriptors worked as long as the change in viewing angle was minimal. A couple of recent works~\cite{oliveira2021sonarOrb, Tueller2018}  give a summary of different keypoint detectors and feature extractors on sonar images and they along with~\cite{Westman18icra} suggest the need for an invariant sonar specific feature descriptor. 


Hand-designing a new feature descriptor for a specific type of sensor is a viable approach~\cite{hansen2007scale, YANG2016_3DLD}, and there has been recent development on a SIFT-like descriptor made for multi-beam sonar~\cite{zhangSIFT}.
The drawback to this process is that the descriptor parameters would need to be manually tuned for different imaging sonar models, as each sonar make has a unique elevation, bearing and range specification, and signal-to-noise profile. On the other hand, recent research on learned feature descriptors for cameras~\cite{detone2018superpoint,sun2021loftr,gleize2023silk} and 3D lidars~\cite{Dewan20183DLD} have shown encouraging results when used for correspondence estimation. Similarly, methods like \cite{triplet2018sonar} leveraged deep neural networks for place recognition for sonar images but have not learned feature correspondences.
These methods utilized deep networks to solve the problem of large variations across scenes, and multiple sensor makes. Looking specifically towards research for camera images, there have been several supervised methods such as SuperPoint~\cite{detone2018superpoint} and LOFTR~\cite{sun2021loftr}. SuperPoint's network works on producing the learned feature descriptors using homographic adaptations for supervision. Follow-up work in SuperGlue, and more recently, LightGlue~\cite{sarlin2020superglue, lindenberger2023lightglue} utilize the SuperPoint feature descriptors to form a robust feature matching framework. Approaches similar to SuperPoint rely heavily on abundant training data, for example from the MegaDepth dataset~\cite{MegaDepthLi18}. Other approaches, like the current state of the art, SiLK~\cite{gleize2023silk}, depend on sub-pixel ground truth correspondences as a linear mapping during training. 

These techniques, while likely to yield similar great results for sonar, are a challenging endeavor to replicate for imaging sonar. This is primarily owing to the scarcity of readily available open-access sonar data and a dearth of ground truth feature point and correspondence information. Additionally, for sonar images, the motion model for planar scenes do not reduce to a homography~\cite{Negahdaripour13tro}.  To address these issues, we look towards ongoing research dedicated to semi-supervised approaches that harness sensor pose data. This enables us to circumvent the necessity for precise ground truth correspondences among feature points. These methods have demonstrated success in achieving equivalent, if not better accuracy compared to their fully supervised counterparts, all while demanding a smaller volume of training data. A noteworthy example of a pose supervised method is CAPS~\cite{wang2020learning}. With access to pose information relating to two images capturing the same scene, CAPS effectively employs a pair of loss functions: epipolar loss and cyclic loss. The foundation of their system rests upon the fundamental notion that a point of interest in the initial image should invariably align with the epipolar line corresponding to its counterpart in the second image. Our approach builds on the backbone of CAPS, utilizing an analog to epipolar geometry for imaging sonars which is detailed in Section~\ref{sec:System}. This gives a data and time-efficient way of obtaining trained feature descriptors for sonar images which can outperform the currently popular methods in use.

\section{Preliminaries}
\subsection{Imaging Sonar Sensor Model\label{sec:Imaging-Sonar-Sensor}}

\begin{figure}
\centering
\includegraphics[width=0.8\columnwidth]{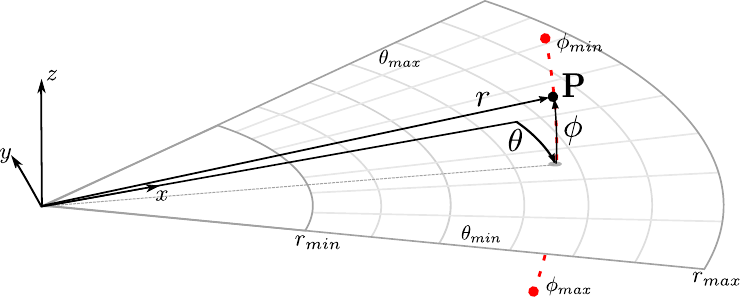}\caption{The basic imaging sonar sensor model of a point feature. Each pixel provides direct measurements of the bearing / azimuth ($\theta$) and range ($r$), but the elevation angle ($\phi$) is lost in the projection onto the image plane - analogous to the loss of the range in the perspective projection of a camera. The imaged volume, called the frustum, is defined by the sensors limits in azimuth $\left[\theta_{min},\theta_{max}\right]$, range $\left[r_{min},r_{max}\right]$, and elevation $\left[\phi_{min},\phi_{max}\right]$.
}\vspace{-0.3cm}
\label{fig:sensor-model-foundations}

\end{figure}

Imaging sonars are active acoustic sensors that measure reflection intensities of emitted sound waves. While analogous to optical cameras in projecting 3D scenes to 2D, their output images are distinctly different. 
We use the following coordinate frame convention, the $x$ axis points forward from the acoustic center of the sensor, with the $y$ axis pointed to the left and $z$ axis pointed up, as shown in Fig.~\ref{fig:sensor-model-foundations}. Unlike the pinhole camera model, where the scene from the viewable frustum is projected onto a forward-facing image plane, for imaging sonars the scene is projected into the zero elevation plane, which is the $xy$ plane in the sonar frame.

Modifying the notation in \cite{Hurtos14jfr} slightly, consider a point $\mathbf{P}$ with spherical coordinates $\left(r,\theta,\phi\right)$ - range, azimuth, and elevation, with respect to the sonar sensor. The corresponding Cartesian coordinates are then represented as:
\begin{equation}
\mathbf{P}=\left[\begin{array}{c}
X_{s}\\
Y_{s}\\
Z_{s}
\end{array}\right]=r\left[\begin{array}{c}
\cos\theta\cos\phi\\
\sin\theta\cos\phi\\
-\sin\phi
\end{array}\right]\label{eq:polar-to-cartesian}
\end{equation}
In the pinhole camera model, a pixel location can indicate azimuth and elevation angles, but lacks clear range information. Any 3D point along its ray projects to the same pixel. Conversely, in the imaging sonar model, a 3D point is mapped onto the zero elevation plane as
\begin{equation}
\mathbf{p}=\left[\begin{array}{c}
x_{s}\\
y_{s}
\end{array}\right]=r\left[\begin{array}{c}
\cos\theta\\
\sin\theta
\end{array}\right]=\frac{1}{\cos\phi}\left[\begin{array}{c}
X_{s}\\
Y_{s}
\end{array}\right].
\end{equation}

In the projective camera model, each pixel has a corresponding ray that passes through the sensor origin. In contrast, the sonar sensor model has a finite elevation arc in 3D space, as seen in Fig.~\ref{fig:sensor-model-foundations} as a red dotted line. In sonar images, a pixel conveys azimuth and range data for an arc but lacks elevation, similar to the range data being absent in the projective camera model. This nonlinear projection and limited field of view in imaging sonar sensors create complexity, making tasks that are straightforward for optical cameras seem challenging for sonar. Another complication arises from the information these pixels hold. In camera images, pixels represent light intensity from specific surface patches along corresponding rays, often resulting in unique 
pixel-to-patch correspondence. This similarity in pixel values across viewpoints accommodates minor camera origin shifts. In contrast, acoustic images may have pixels representing multiple surfaces along an elevation arc reflecting sensor-emitted sound. This can lead to compounded intensity in a single pixel, which may vary even with slight sensor origin changes.

\subsection{Sonar Epipolar Geometry}~\label{subsec:Sonar-Epipolar}

Negahdaripour introduced the concept of stereo epipolar geometry in~\cite{Negahdaripour18oceans}. In cameras, the epipolar line is the intersection of a point's epipolar plane with the image plane. Essentially, it's the projection of the line connecting the 3D point and one camera center onto the second camera's image plane. This line represents the depth and direction from the first view. In sonar stereo, elevation arcs serve as the analog to the epipolar lines found in cameras, and their projection is an \emph{epipolar contour}. We will describe the epipolar geometry in brief here, but refer the reader to~\cite{Negahdaripour18oceans} for a detailed explanation. 

We refer to Section \ref{sec:Imaging-Sonar-Sensor} where we return to Eq.~\ref{eq:polar-to-cartesian} for notation relating to the conversion of a point in polar space to the Cartesian space. Pixels are in range-bearing format since we are training sonar images. Given a point in one sonar image at some range, $R$ and bearing $\theta$. The point in 3D will lie along the elevation arc at the same range and bearing as defined and at some arbitrary elevation angle, $\varphi$. Since we do not know the elevation angle, the ambiguity is along the elevation arc for $\varphi_{min}<=\varphi<=\varphi_{max}$. Now, the conversion of points in 3D Cartesian space to the range bearing space is as seen in Eq.~\ref{eq:cartesian-to-polar}. 
\begin{equation}
\boldsymbol{P}=\left[\begin{array}{c}
R\\
\theta\\
\phi
\end{array}\right]=r\left[\begin{array}{c}
\sqrt{x^{2}+y^{2}+z^{2}}\\
arctan2(y,x)\\
arctan2(x,\sqrt{x^{2}+y^{2}})
\end{array}\right]\label{eq:cartesian-to-polar}
\end{equation}

\begin{figure}
\centering{}\includegraphics[width=0.8\columnwidth]{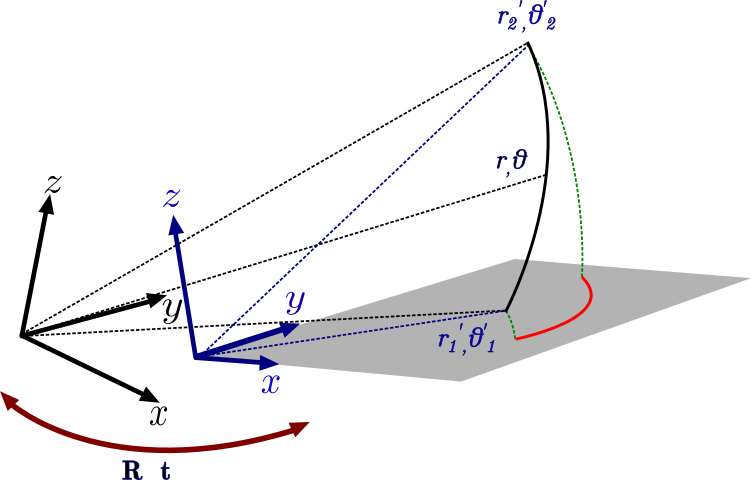}\caption{Sonar Epipolar Geometry: The elevation arc of a point in the first
image is transformed into the frame of the second image and then projected,
which creates an epipolar contour.\vspace{-0.3cm} \label{Fig: Sonar-epipolar}}

\end{figure}

Assuming we have a feature point $\boldsymbol{p}$ in the first image, and $\boldsymbol{R_{1,2}}$and \textbf{$\boldsymbol{t_{1,2}}$} are the known rotation and translation between the coordinate frame of image 1 and image 2. The locus of the same feature point in image 2, $\boldsymbol{p^{{\textstyle '}}}$ is calculated as seen in Eq.~\ref{eq:feature-locus}. This 3D feature locus can now be projected into the polar sonar image plane as in Eq.~\ref{eq:project-to-polar}. Fig.~\ref{Fig: Sonar-epipolar} visually describes the process of projecting the elevation arc from the first frame onto the second image plane. The red line is the epipolar contour we use for the loss function described in the following section. 
\begin{equation}
\boldsymbol{R_{1,2}}=\left[\begin{array}{c}
r_{1}\\
r_{2}\\
r_{3}
\end{array}\right],\boldsymbol{t_{1,2}}=[\begin{array}{ccc}
t_{x} & t_{y} & t_{z}\end{array}]^{T}
\end{equation}
\begin{equation}
\boldsymbol{p}'(\phi)=\left[\begin{array}{c}
x'\\
y'\\
z'
\end{array}\right]=\left[\begin{array}{c}
r_{1}\cdot P+t_{x}\\
r_{2}\cdot P+t_{y}\\
r_{3}\cdot P+t_{z}
\end{array}\right]\label{eq:feature-locus}
\end{equation}
\begin{equation}
\boldsymbol{p_{proj}^{{\textstyle '}}}(\phi)=\left[\begin{array}{c}
r'\\
\theta'
\end{array}\right]=\left[\begin{array}{c}
\left\Vert P'_{p}\right\Vert _{2}\\
arctan2(y',x')
\end{array}\right]\label{eq:project-to-polar}
\end{equation}

\section{Method\label{sec:System}}

In this section we describe the key aspects of our method. We first discuss obtaining keypoints, then introduce the network, and detail the loss functions and the training strategy. 

\subsection{Keypoint Detection}
SONIC, like CAPS, trains feature representations and requires keypoints as an input. We use a combination of AKAZE and SuperPoint keypoints for training to ensure we have enough keypoints per frame while training. Keypoints are found in polar space as converting sonar images to euclidean spaces leads to loss of information, especially when near the sonar.

\subsection{Network Overview}

\begin{figure}
\centering{}\subfloat[Differentiable Matching Layer]{\includegraphics[width=0.8\columnwidth]{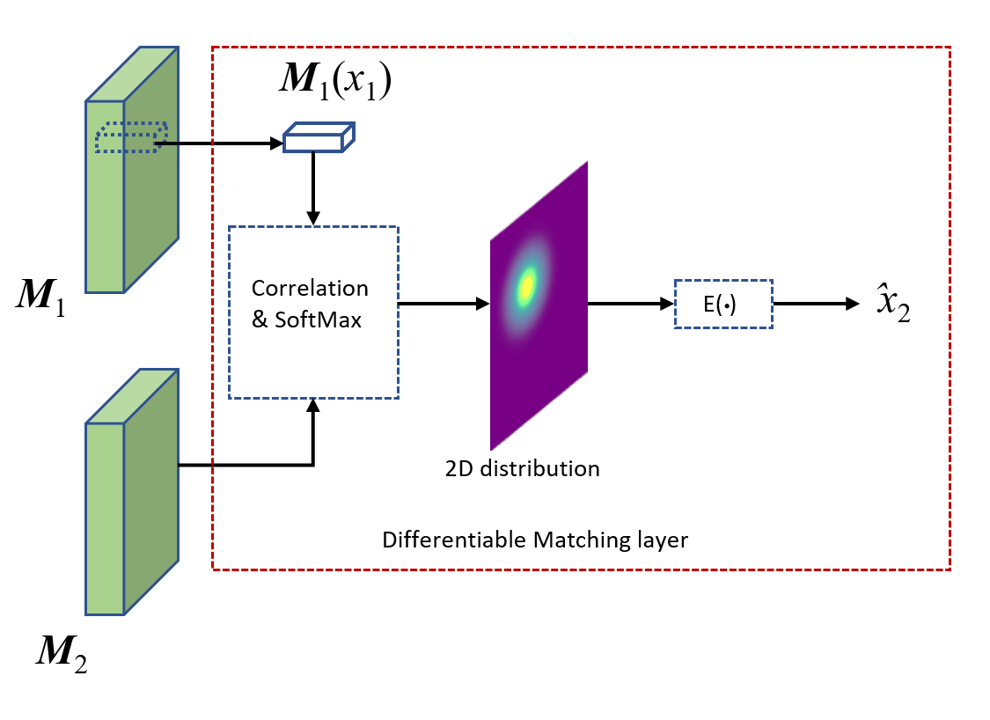}}

\centering{}\subfloat[Coarse to Fine Module]{\includegraphics[width=0.8\columnwidth]{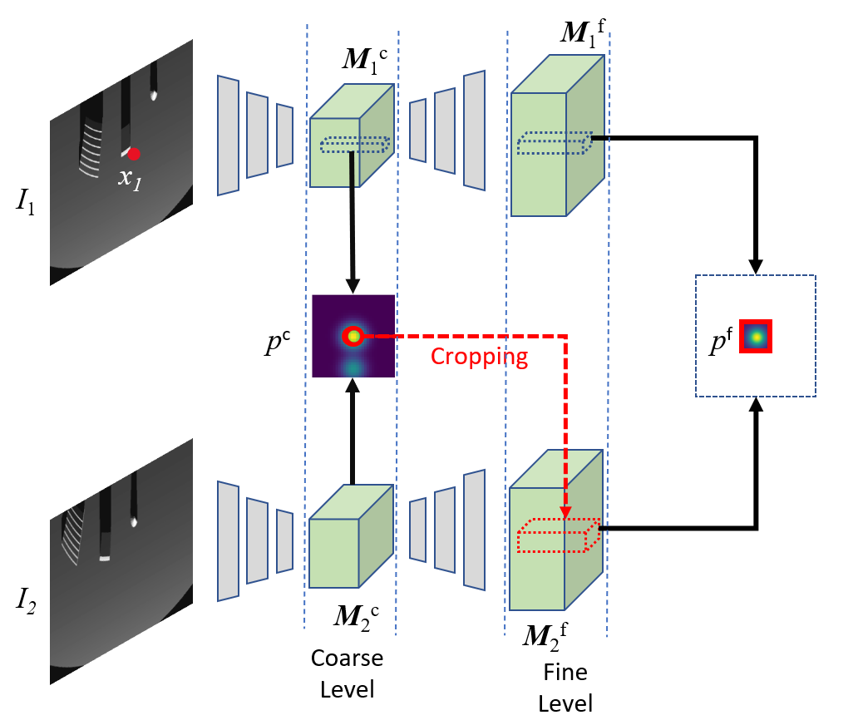}

}\caption{Network architecture highlights: a) For each qeury point $x_{1}$, its corresponding location $\hat{x}_{2}$(Eq.~\ref{eq:expected_corerespondence}) is represented as the expectation of a distribution computed from the correlation between the feature descriptors. The associated uncertainty also helps in reweighting training loss. During training, keypoints serve as queries (b) Searching correspondence across the entire image is costly. The location of the correspondence $p^{c}$ at the coarse level is used to ascertain a local window at the fine level, $p^{f}$ is found in this window using differentiable matching. \vspace{-0.5cm}
\label{fig:Network-Architecture}}
\end{figure}

We use a CNN based encoder-decoder architecture with a differentiable matching layer and coarse-to-fine technique similar to~\cite{wang2020learning}. Unlike CAPS, SONIC uses a ResNet-34~\cite{he2015resnet}  base to prevent overfitting our small dataset.  
From here, similar to CAPS, the encoder creates a coarse representation; given the input keypoint we find the corresponding point in this representation using the differentiable matching layer. A pictorial representation of the differentiable matching layer and coarse to fine architecture is shown in Fig.~\ref{fig:Network-Architecture} (a) and (b) respectively. Given two images, the network with shared weights creates the representation $M_{1}$ and $M_{2}$. For each query point $x_{1}$, the feature descriptor $M_{1}(x_{1})$ is correlated to each point in $M_{2}$. This is used to find the probability of each point being the correspondence of $x_{1}$ in $M_{1}$ as shown in Eq.~\ref{eq:x1_probability}. A correspondence is calculated by finding the expectation of this probability in Eq.~\ref{eq:expected_corerespondence}. 
Searching correspondences for all points over the image is very computationally costly, hence it makes most sense to sparsely sample the query points for supervision. This coarse to fine architecture improves on efficiency. At the coarse level, a correspondence distribution is computed over all the locations. However, at the finer level, the distribution is only computed at the highest probability location observed from the coarse map. The loss functions are imposed on both levels.

Our other modifications to the original CAPS architecture include changes to work with single channel sonar images, and condense final coarse and fine layer outputs to 64 dimensions instead of the original 128.   
\begin{equation}
p(x|x_1, M_1, M_2) = \frac{\exp (M_1(x_1)^\top M_2(x))}{\sum_{y \in I_2} \exp (M_1(x_1)^\top M_2(y))}
\label{eq:x1_probability}
\end{equation}
\begin{equation}
\hat{x}_2 = h_{1 \rightarrow 2}(x_1) = \sum_{x \in I_2} x \cdot p(x|x_1, M_1, M_2)
\label{eq:expected_corerespondence}
\end{equation}

\subsection{Supervision}

\begin{figure}
\centering{}\includegraphics[width=1\columnwidth]{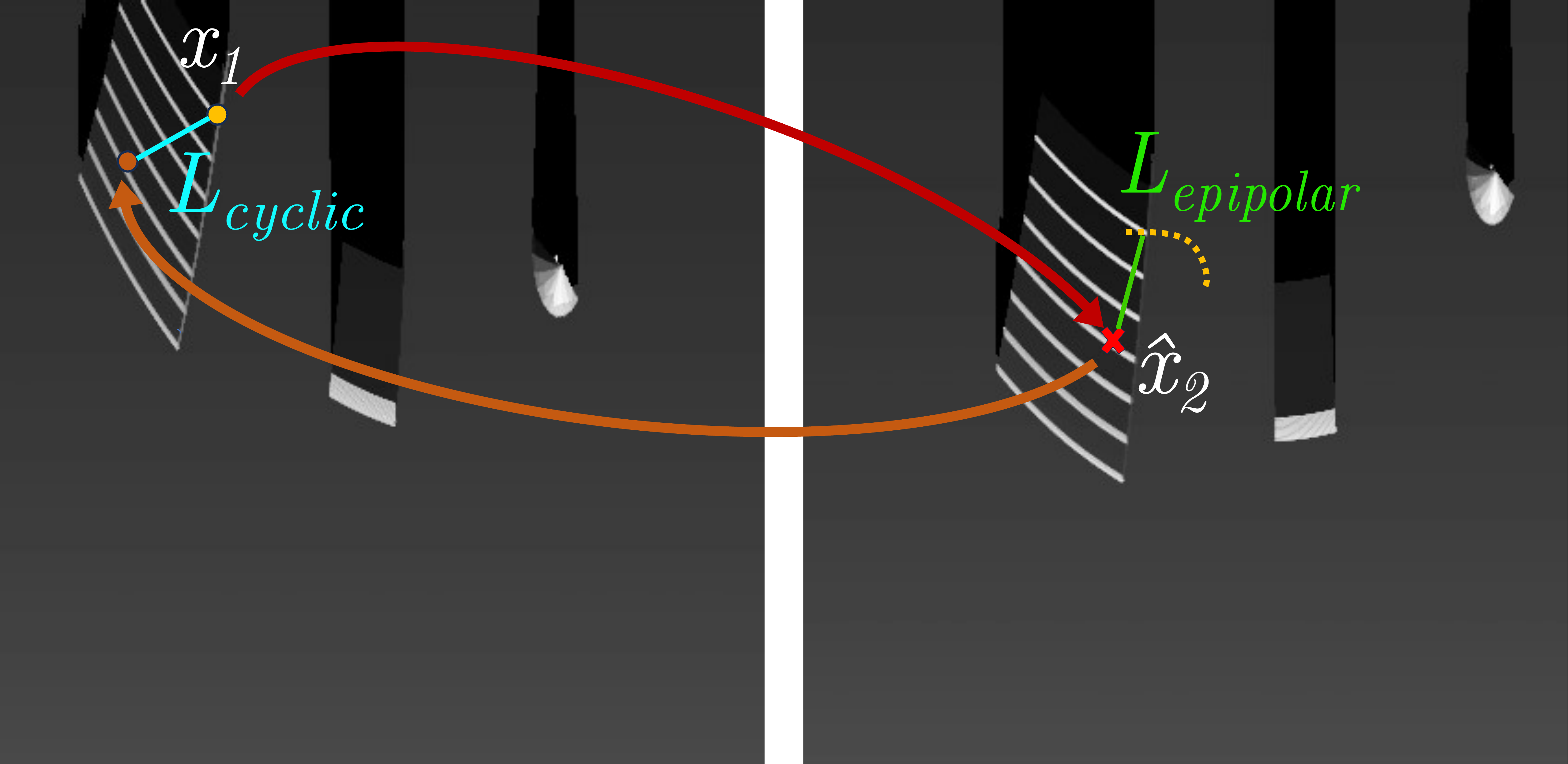}\caption{Loss functions: The yellow point $x_{1}$ represents a queried keypoint
in the first image. The red cross $\hat{x_{2}}$is the predicted point. The yellow dotted line represents the sampled points on the epipolar contour of point $x_{1}$. 
$L_{epipolar
}$ is the shortest distance to the epipolar contour, or simply
the epipolar loss. $L_{cyclic}$ is the cyclic loss to assert that the
mapping of the feature point is close to its original position.\vspace{-0.4cm}~\label{Fig: Sonar-losses}}
\end{figure}

Like CAPS, we propose two loss functions, modifying it for sonar images. We introduce the sonar-epipolar and sonar-cyclic loss. Given the relative pose between two image frames $I_{1}$and $I_{2}$, we use Eq.~\ref{eq:feature-locus} and~\ref{eq:project-to-polar} to determine the epipolar contour for a given point $P$ in the $I_1$. A correctly predicted point in the $I_2$ would lie on this contour. We use the distance between this predicted point and contour as a metric to optimize over. The epipolar loss term, $L_{epipolar}$ for a given point $x_1$  in $I_1$ in Eq.~\ref{eq:epipolar_loss}  is defined as the shortest distance between the predicted correspondence of $\hat{x}_2$ and the epipolar contour of $x_{1}$ in the second image $I_{2}$. In our implementation, we sample points along the elevation arc of the first point in the first image and then transform and project them on to the second image to create a discrete epipolar contour of the sampled points. In the polar frame, the loss is thus the minimum of the distance between the predicted point and each point on the arc as seen in Eq.~\ref{eq:polar_epipolar_loss}. The epipolar loss only checks for the predicted match to lie on the estimated contour. To further constrain the system, a cyclic consistency loss,  $L_{cyclic}$, is utilized which aims to keep the forward-backward mapping of the point consistent. The weighted sum of the losses $L_{epipolar}$ and $L_{cyclic}$ is our final loss function as seen in Eq.~\ref{eq:polar_epipolar_loss}. A point to note is that the distances found for both the losses are in the range-bearing space. Due to nature of sonar images, it is important for the network to learn this distinction, and the combined loss terms in this representation support this. A graphical representation of the losses is seen in Fig.~\ref{Fig: Sonar-losses}. 
\begin{equation}
L_{epipolar}(x_{1})=dist(h_{1\rightarrow2}(x_{1}),ep\_contour)\label{eq:epipolar_loss}
\end{equation}
\begin{equation}
L_{cyclic}(x_{1})=\left\Vert h_{2\rightarrow1}(h_{1\rightarrow2}(x_{1})-x_{1})\right\Vert _{2}\label{eq:cyclic-loss}
\end{equation}
\begin{equation}
L_{(I_{1},I_{2})}=\mathop{\sum_{i=1}^{n}}[L_{epipolar}(x_{1}^{i})+\lambda L_{cyclic}(x_{1}^{i})]\label{eq:joint-loss}
\end{equation}
\begin{equation}
    L_{cyclic}^{1,2}=r_{1}^{2}+r_{2}^{2}-2r_{1}r_{2}(cos(\theta_{1}-\theta_{2}))
\end{equation}
\begin{multline}
L_{epipolar}(p^\prime_{proj}(\varphi),r_{2},\theta_{2})=\\\mathop{\min_{\phi}}(r^\prime(\varphi)^{2}+r_{2}^{2}-2r^\prime(\varphi)r_{2}(cos(\theta^\prime(\varphi)-\theta_{2})))\label{eq:polar_epipolar_loss}
\end{multline}

\subsection{Data and Training}

Acquiring numerous sonar images with ground truth pose is challenging, and real-world data often lacks feature-rich frames for network training. While there are sonar datasets like those by Singh et al. \cite{Singh2021Sonar} and Malios et al. \cite{Mallios2017ijrr} for object classification or SLAM, they either lack pose information or do not use wide elevation imaging sonars. Thus, simulated environments, particularly the HoloOcean simulator \cite{potokar2022holoocean,Potokar2022iros}, supplemented by custom worlds, offer an enhanced solution for data generation.

For our primary application, seafloor mapping, sonar images were gathered from 1-4m above the floor at a downward pitch of 10-20$^\circ$. Multiple trajectories with varying rotation and translation offsets produced image pairs. Our dataset comprises approximately 300K training and 30K validation pairs. We configure our simulated images on the Blueprint Subsea M1200d's~\cite{Blueprint} low frequency mode. In this mode, the images have a maximum azimuthal field of view of 130$^\circ$ and elevation of 20$^\circ$. The maximum range is set to be 10m. The image is comprised of 512$\times$512 range and bearing bins. Image noise parameters are marginally varied around the values provided in HoloOcean's example scenarios.

Epipolar and cyclic losses are given 0.7 and 0.3 weights, respectively. We train the network for $\approx7$ epochs with a batch size of 14 pairs, using a Nvidia GeForce 4090 RTX with 24GB memory.

\section{Evaluation}

We assessed matching performance by comparing SONIC, AKAZE, and the camera-trained LightGlue model. As highlighted in Section~\ref{sec:Related} AKAZE is the preferred keypoint and descriptor for imaging sonar applications. LightGlue~\cite{lindenberger2023lightglue} is a recent improvement on SuperGlue, with a performance equivalent to LoFTR~\cite{sun2021loftr} and MatchFormer~\cite{wang2022matchformer} which are considered the current state of the art for camera image correspondence. We compare SONIC with LightGlue since both use SuperPoint keypoints and require a keypoint detector, unlike other detector-free matchers. 

Assuming a planar scene we evaluate the number of matched keypoints that are considered to be inliers for each of AKAZE, LightGlue and SONIC. Inliers are classified by projecting the keypoints from the reference image using the ground truth relative pose information as described in Section~\ref{subsec:Sonar-Epipolar} onto the query image. The threshold selected for simulation images is 12 pixels, which corresponds to $0.23$m in range or $3^\circ$ in bearing. We increase this threshold for real images to be 20 pixels, which corresponds to $0.2$m and $5^\circ$ in range and bearing respectively. 
To show the benefit of our method towards downstream tasks such as feature-based SLAM, we evaluate performance in relative sensor pose recovery using the two-view acoustic bundle adjustment framework to recover sensor pose information as presented by Westman et al.~\cite{Westman19joe}.
It is typical for hovering underwater vehicles to be equipped with pressure sensors for depth estimation, and to be limited to planar motion~\cite{Kaess10istmp}.  Most graph-based solutions for underwater SLAM would thus model \textit{Z}, \textit{pitch} and \textit{roll} as a separate unary factor, independent to planar motions in \textit{X}, \textit{Y}, and \textit{Yaw}~\cite{Teixeira16iros}. Thus  we focus mainly on evaluating the average absolute error in planar translation (\textit{xy}) and rotation (\textit{yaw}). 
Prior pose estimates are derived from corrupted ground truth data. These poses are used to filter matches.
We perform a Z-test on the distance between the corresponding projected and matched query keypoints. We prune matches not falling under a $2\sigma$ threshold for all three methods.  The same noisy pose prior is provided to the acoustic bundle adjustment optimizer along with matched keypoints from the reference and query images, passed as their bearing and range values. In SLAM application, the data association could be improved by using RANSAC~\cite{ransacOceansKim} or JCBB~\cite{Westman19joe}.

We present and discuss results for our simulated datasets first, and then for the real world data from a test tank.

\subsection{Simulation}
\begin{figure}

\includegraphics[width=1\columnwidth]{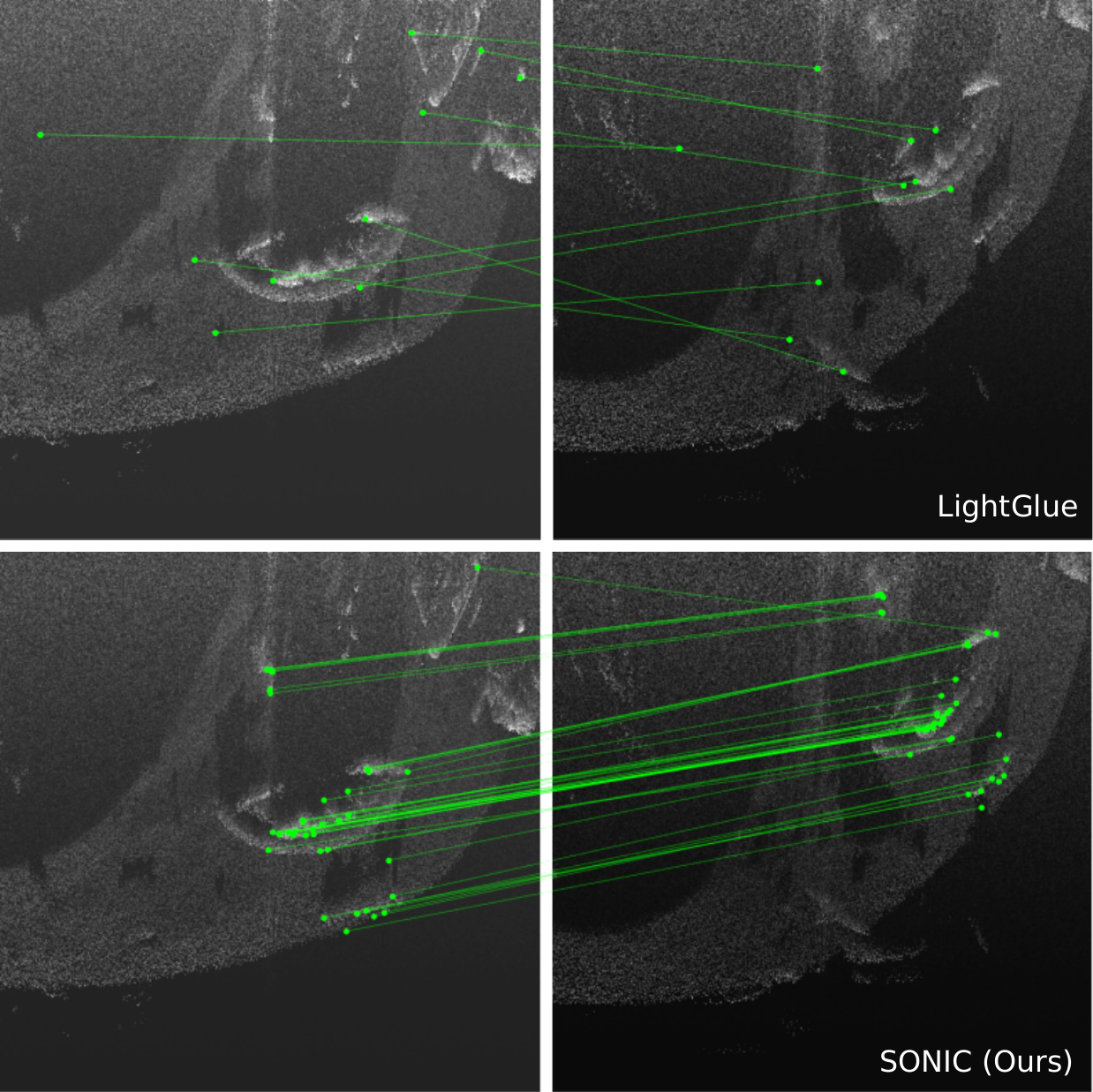}\caption{Simulation Matching Performance: LightGlue is unable to match features when the structures' shape warps under sensor motion, a common phenomenon in polar images. \vspace{-0.3cm}}\label{fig:Simulation_comparison}
\end{figure}

We sample sonar image pairs, unused for SONIC's training, with varied pose differences including minor roll, pitch, and z variations, and broader x, y, and yaw differences. These pairs are categorized into two groups: small and large variation. The small variation group contains pairs with \textit{yaw} $<$ $\pm5^\circ$ and \textit{x,y} $ < \pm1.5$m. The large variation group contains pairs with up to $\pm40^\circ$ variation in rotation and $\pm7$ meter in translation. Evaluation images had the same parameters for range, bearing and elevation as used in training. 
We first look at the inlier ratio in ~\ref{tab:Inlier_results}, where both small and large variation groups show significantly more ground truth pose agreeable matches than the other two methods. We then look at Table~\ref{Tab:Results} to see the average absolute error and standard deviation for planar translation(meters) and rotation(radians). Our method outperforms the other two in all situations. SONIC is specifically trained to identify features which are more likely to be invariant to viewpoint changes, as well as learn the geometric variance of features across these changes. The inductive bias of convolutional neural nets also ensures that it looks at a area and structure around the keypoint and finds matches closest to that in the other image. AKAZE and other traditional descriptors try to describe features such as as corners, edges and blobs which do not stay consistent in sonar images.  LightGlue, and other learned matchers also look at the image as a whole and thus perform much better than AKAZE. However, since LightGlue is trained on camera images it cannot predict accurate matches when the structures or keypoints change over significant translation or rotational variation due to the change in object geometries occurring in the polar image-space as seen in Fig.~\ref{fig:Simulation_comparison}. 

\subsection{Test Tank Evaluation}

Transferring simulated performance to real-world scenarios is challenging. As SONIC was solely trained on simulated data, it's crucial to assess if it learned the geometry of view-invariant features in actual sonar data. Accurately modeling sonar noise is challenging, making real-image performance evaluation vital.
We conducted experiments in a test tank with a sliding gantry system with a configurable 6 degrees of freedom sensor pose as seen in Fig.~\ref{fig:match_performance}. We used a Leica Total Station 16~\cite{leicageosystems} with a tracking prism for ground truth estimation. Here, we show two sample frames, taken  at differing positions with a change of $1.153$m and $1.310$m in $\mathbf{x}$ and $\mathbf{y}$ and a yaw rotation of $-15^\circ$. The other degrees of freedom remained constant.  We observe our method is able to provide the right matches, whereas previously used methods like AKAZE with symmetric nearest-neighbour brute force matching and SuperPoint descriptors with LightGlue are unable to do so.  The matching results of the three methods in the tank can also been seen in the same figure. While we trained our images on a maximum range setting of  10 meter, the real world  parameters was fixed to $6$m due to the small size of the tank ($7$m diameter). We also had to limit the keypoint detectors for all three methods from detecting points beyond the general position of the objects due to significant acoustic reflections between the metallic surfaces of the tank wall and pipe placed inside the tank. 
Due to the limited change in pose, and concentrated features, the two-view acoustic bundle adjustment framework was unable to resolve the relative pose for most image pairs. However, we can see from the inlier ratio in Table~\ref{tab:Inlier_results} that SONIC is once again more likely to provide better correspondences.

\newcolumntype{Y}{>{\centering\arraybackslash}X}

\begin{table}[htbp]
    \centering
\caption{Percentage of Inliers}
\label{tab:Inlier_results}
    \begin{tabular}{|c|c|c|c|}
        \hline
        Method & \multicolumn{1}{|p{2cm}|}{\centering Simulation - \\ Small Variation}& \multicolumn{1}{|p{2cm}|}{\centering Simulation - \\ Large Variation} & {\centering Test Tank}\\
        \hline
        AKAZE& 24.23\%& 10.22\%& 40.08\%\\
        \hline
        LightGlue& 39.13\%& 11.18\%& 51.92\%\\
        \hline
        \textbf{SONIC}& \textbf{49.43}\%& \textbf{23.56}\%& \textbf{74.53}\%\\
        \hline
    \end{tabular}

\end{table}

\begin{table}[htbp]
    \caption{Planar Translation and Rotational Accuracy}
     \label{Tab:Results}
    \centering
    \begin{tabularx}{\linewidth}{|c|Y|Y|Y|Y|}
        \hline
        \multirow{3}{*}{Method} & \multicolumn{4}{c|}{Simulated data} \\
        \cline{2-5}
        & \multicolumn{2}{c|}{Small Variation} & \multicolumn{2}{c|}{Large Variation} \\
        \cline{2-5}
        & Translation (m) & Rotation (rad) & Translation (m) & Rotation (rad) \\
        \hline
        AKAZE & 
        $\mu$: 4.24,\newline $\sigma$: 2.40 & 
        $\mu$: 1.56,\newline $\sigma$: 1.22 & 
        $\mu$: 5.06,\newline $\sigma$: 2.35 & 
        $\mu$: 1.44,\newline $\sigma$: 1.01 \\
        \hline
        LightGlue & 
        $\mu$: 3.62,\newline  $\sigma$: 4.52 & 
        $\mu$: 0.97,\newline  $\sigma$: 1.12 & 
        $\mu$: 3.49,\newline  $\sigma$: 3.53 & 
        $\mu$: 0.97,\newline  $\sigma$: 1.00 \\
        \hline
        \textbf{SONIC} & 
        $\mu$: \textbf{0.88},\newline  $\sigma$: \textbf{2.01} & 
        $\mu$: \textbf{0.25},\newline  $\sigma$: \textbf{0.61} & 
        $\mu$: \textbf{2.23},\newline  $\sigma$: \textbf{3.35} & 
        $\mu$: \textbf{0.60},\newline  $\sigma$: \textbf{0.82} \\
        \hline
    \end{tabularx}
   
\end{table}

\section{Conclusions and Future Work}

We propose SONIC, a pose-supervised model that solves the challenging problem of feature correspondence in sonar images. Our results demonstrate that SONIC excels over existing techniques in sonar image feature extraction, addressing the need for sonar-centric feature correspondence.  Our method achieves this through a novel sonar epipolar loss and a cyclic consistency loss.  The epipolar loss utlizes the epipolar contour projected by an arc onto the second image using relative pose information. This guides the predicted correspondence to align with this contour. Simultaneously, our consistency loss ensures cyclic congruence between the predicted and the initial keypoints.
Our method, trained solely on simulated data, demonstrates promising real-world feature correspondence due to its understanding of underlying geometry. However, more thorough open water tests are needed, along with supplementing real data towards training to close the gap between simulation and real-world results. 
Future applications of this work include the extension of the method to incorporate different sonar frequency modes and parameters under one model. Recent work on image matching using attention~\cite{wiles2021co} opens the door to enable cross-sonar feature matching and improving performance, which could potentially help cross-platform localization and mapping.

{\footnotesize\bibliographystyle{ieeetr}
\bibliography{references}}

\end{document}